\documentclass[conference]{IEEEtran}
\IEEEoverridecommandlockouts
\usepackage{cite}
\usepackage{amsmath,amssymb,amsfonts}
\usepackage{algorithmic}
\usepackage{comment}
\usepackage{tabularx}
\usepackage{multirow}
\usepackage{bm}
\usepackage{booktabs}
\usepackage{graphicx}
\usepackage{textcomp}
\usepackage{xcolor}
\usepackage{balance}

\usepackage{hyperref}

\usepackage{times}
\usepackage{latexsym}
\usepackage{amssymb}
\usepackage{amsthm}
\usepackage{multicol}
\usepackage{amsmath}
\usepackage{multirow}
\usepackage{enumitem}
\usepackage{booktabs}
\usepackage{array}
\usepackage{bm}
\usepackage{graphicx}
\usepackage{pbox}
\usepackage{adjustbox}
\usepackage{colortbl}

\def\BibTeX{{\rm B\kern-.05em{\sc i\kern-.025em b}\kern-.08em
    T\kern-.1667em\lower.7ex\hbox{E}\kern-.125emX}}
\begin{document}

\title{Prompting Large Language Models for \\ Topic Modeling}



\author{\IEEEauthorblockN{Han Wang\IEEEauthorrefmark{1}, Nirmalendu Prakash\IEEEauthorrefmark{1}, Nguyen Khoi Hoang\IEEEauthorrefmark{2}, Ming Shan Hee\IEEEauthorrefmark{1}, Usman Naseem\IEEEauthorrefmark{3}, Roy Ka-Wei Lee\IEEEauthorrefmark{1}}
\IEEEauthorblockA{Singapore University of Design and Technology\IEEEauthorrefmark{1}, Singapore\\ 
VinUniversity\IEEEauthorrefmark{2}, Vietnam\\
James Cook University of North Queensland\IEEEauthorrefmark{3}, Australia\\
Email: han\_wang@sutd.edu.sg, nirmalendu\_prakash@sutd.edu.sg, 20nguyen.hk@vinuni.edu.vn, \\mingshan\_hee@mymail.sutd.edu.sg, usman.naseem@jcu.edu.au, roy\_lee@sutd.edu.sg}}

\makeatletter
\def\ps@headings{
\let\@oddhead\@empty
\let\@evenhead\@empty
\def\@oddfoot{\@IEEEheaderstyle\hfil\thepage}%
\def\@evenfoot{\@IEEEheaderstyle\thepage\hfil\hbox{}}
a}
\def\ps@IEEEtitlepagestyle{
\let\@oddhead\@empty
\let\@evenhead\@empty
\def\@oddfoot{\footnotesize 979-8-3503-2445-7/23/\$31.00 ©2023 IEEE\hfill\thepage}%
\let\@evenfoot\@empty
}
\makeatother

\maketitle

\begin{abstract}
Topic modeling is a widely used technique for revealing underlying thematic structures within textual data. However, existing models have certain limitations, particularly when dealing with short text datasets that lack co-occurring words. Moreover, these models often neglect sentence-level semantics, focusing primarily on token-level semantics. 
In this paper, we propose \textsf{PromptTopic}, a novel topic modeling approach that harnesses the advanced language understanding of large language models (LLMs) to address these challenges. It involves extracting topics at the sentence level from individual documents, then aggregating and condensing these topics into a predefined quantity, ultimately providing coherent topics for texts of varying lengths. This approach eliminates the need for manual parameter tuning and improves the quality of extracted topics. 
We benchmark \textsf{PromptTopic} against the state-of-the-art baselines on three vastly diverse datasets, establishing its proficiency in discovering meaningful topics. Furthermore, qualitative analysis showcases \textsf{PromptTopic}'s ability to uncover relevant topics in multiple datasets.
\end{abstract}

\section{Introduction}
\textbf{Motivation}. Topic modeling stands as a pivotal statistical method, focused on discerning latent thematic patterns in textual datasets~\cite{blei2003latent}. It has secured a substantial footing in diverse areas, such as information retrieval and text mining. Its prowess in efficiently extracting topics from voluminous document collections bestows researchers with the capacity to delve into vast textual datasets in an efficient manner.

Over the years, topic modeling has burgeoned as an integral domain within natural language processing and machine learning. Pioneering endeavors in the field leaned towards a bag-of-words probabilistic framework to ascertain topics~\cite{blei2003latent,10.5555/1036843.1036902, blei2006correlated,fevotte2011algorithms}. Contemporary strides, however, have steered towards embracing word embedding-centric~\cite{xun2017correlated,dieng2020etm}, neural frameworks~\cite{srivastava2017autoencoding}, and transformer paradigms~\cite{bianchi2020cross,grootendorst2022bertopic}—all in a bid to encapsulate subtler intricacies within textual compositions.

Yet, the evolution hasn't rendered the field impervious to challenges. Encounters with unfamiliar words still pose significant hurdles, attributed to the fact that these models conventionally thrive on predetermined lexicons. Moreover, the fixation on word-level analysis often overshadows the deeper, contextual essence embedded within sentences. Additionally, the recurrent need to meticulously adjust hyperparameters for superior outputs makes these models not only resource-intensive but also intricate to handle.

\textbf{Research Objectives}. In light of these pressing challenges, our research introduces \textsf{PromptTopic}. This avant-garde, prompt-driven strategy for topic modeling taps into the vast potential of large language models (LLMs). Explicitly, \textsf{PromptTopic} integrates capabilities of renowned LLMs like ChatGPT\footnote{\url{https://api.openai.com/v1/chat/completions}} and LLaMa~\cite{touvron2023llama} to seamlessly intertwine word and sentence semantics—paving the way for a more holistic topic modeling experience. Our method meticulously crafts prompts that are adept at isolating lucid and actionable topics from texts, eliminating the often laborious fine-tuning phase. The adoption of in-context learning through prompts further negates the time-consuming hyperparameter calibration, thus refining the entire topic modeling paradigm.

\textbf{Contributions.} The main contributions of this work are as follows: 1. We propose \textsf{PromptTopic}, a novel prompt-based model to perform topic modeling on text. To the best of our knowledge, this is the first topic modeling model that utilizes LLMs. 2. We conduct comprehensive experiments on three widely used topic modeling datasets to evaluate the performance of \textsf{PromptTopic} compared to state-of-the-art topic models. 3. We conduct a qualitative analysis of the learned topics, highlighting that our model exhibits the ability to identify meaningful and highly coherent topics.

\section{Related Work}
\subsection{Topic Modeling}

Topic modeling, a significant area within natural language processing and information retrieval, is centered on unveiling abstract "topics" within a document collection. Topic modeling also has been applied to many other domain such as identifying topical influential users in social media~\cite{lee2017analyzing,lee2018discovering,lee2019discovering}. Over time, a plethora of methods have evolved to improve the topic modeling performance. A standout model is the Latent Dirichlet Allocation (LDA) by \cite{blei2003latent}, which has been foundational for later innovations. This model was enhanced by its successors, such as supervised LDA (sLDA) \cite{mcauliffe2007supervised} and dynamic topic modeling (DTM) \cite{blei2006dynamic}. Other noteworthy techniques include non-negative matrix factorization (NMF) by \cite{fevotte2011algorithms} and probabilistic latent semantic analysis (pLSA) by \cite{hofmann2013probabilistic}.

The deep learning era brought neural models like the adapted Variational Autoencoder (VAE) and Transformer for topic modeling. ProdLDA by \cite{srivastava2017autoencoding} is one such adaptation of LDA. Recent models like Correlated Topic Model~\cite{xun2017correlated}, Gaussian LDA~\cite{das2015gaussian}, Spherical HDP~\cite{batmanghelich-etal-2016-nonparametric}, and Embedded Topic Model (ETM)\cite{dieng2020etm}, integrate word embeddings like Word2Vec\cite{mikolov2013word2vec} to grasp semantic word relationships. Strategies, including TopClus \cite{meng2022topicdiscovery} and Cluster-Analysis \cite{sia2020tired}, emphasize clustering word and document embeddings, offering flexibility by separating cluster creation from topic representation. Further, models like Contextualized Topic Model (CTM)\cite{bianchi2020cross} and BERTopic\cite{grootendorst2022bertopic} leverage transformers like BERT~\cite{devlin2018bert} to assimilate context and shape topic representations. Similar efforts employ pretrained language embeddings for the same endeavor~\cite{hoyle-etal-2020-improving,gupta-etal-2021-multi}.

In our paper, we introduce \textsf{PromptTopic}, an innovative approach that leverages Large Language Models (LLMs) for topic modeling. Unlike traditional models, it seamlessly incorporates word and sentence semantics, facilitating precise and contextually relevant topic identification. Importantly, 'PromptTopic' streamlines the process by avoiding extensive hyperparameter tuning, making it more accessible to researchers and users.

\begin{figure}[t]
	\centering
	\includegraphics[width=0.45\textwidth]{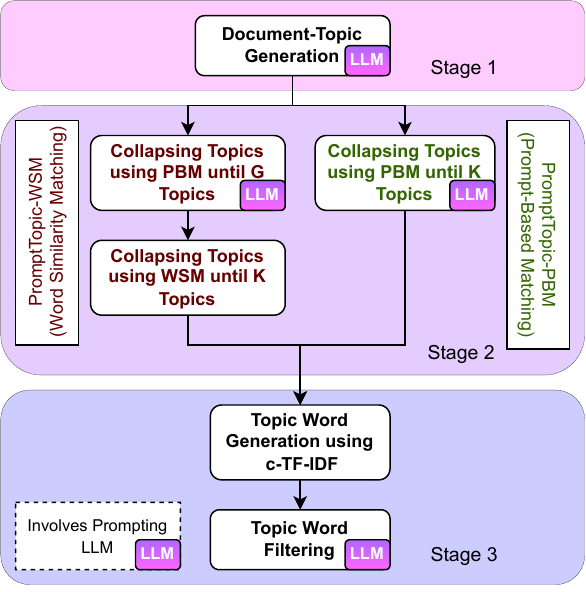} 
	\caption{Overview of \textsf{PromptTopic}.}
	\label{fig:framework}
\end{figure}

\subsection{Large Language Model}

In the dynamic world of large language models (LLMs), foundational works have set the stage for innovative research and applications. The groundbreaking Transformer architecture, introduced by \cite{vaswani2017attention}, laid the groundwork for subsequent LLMs. A milestone was reached in 2020 with the release of GPT-3 by \cite{brown2020language}, a model with a staggering 175 billion parameters, displaying unmatched capabilities. A notable variant of GPT-3, designed for conversations, is ChatGPT.

For a long time, the consensus was that larger models equated to better performance. However, this notion was recently challenged by J. Hoffmann in \cite{hoffmann2022training}. They argued that superior results, within a set budget, can be achieved with smaller models trained on extensive datasets. Touvron demonstrated LLaMA in ~\cite{touvron2023llama}, a set of efficient LLMs with sizes from 7 billion to 65 billion parameters, showcasing their competitiveness against larger LLMs.

The current era of LLMs is marked by multimodality. Models like GPT-4 \cite{openai2023gpt4} and LLaVA \cite{liu2023visualinstruction} now encompass visual processing capabilities, enhancing their versatility.

Our research aims to utilize both accessible online LLM APIs, such as ChatGPT, and offline models like LLaMA, to assess their efficacy in topic modeling scenarios.

\section{Methodology}
The \textsf{PromptTopic} is an unsupervised approach that harnesses the robust language understanding capabilities of LLMs to generate topics. The model consists of three stages: \textit{Topic Generation}, \textit{Topic Collapse,} and \textit{Topic Representation Generation}, as illustrated in Figure \ref{fig:framework}. Each stage leverages LLMs to extract, organize, and refine topics from input documents. Using prompts, the model effectively captures underlying themes and concepts, enhances topic clustering, and improves the quality of topic representations. Prompting LLMs allows \textsf{PromptTopic} to learn document topics comprehensively, eliminating the need for fine-tuning.

\subsection{Topic generation}

\definecolor{question_color}{HTML}{1B9E77}
\definecolor{label_color}{HTML}{D95F02}
\definecolor{generated_gpt}{HTML}{FF0000}
\definecolor{assistant_sample}{HTML}{82B366}
\definecolor{user_input}{HTML}{9673A6}

\begin{figure}[t]
	\centering
	\includegraphics[width=0.45\textwidth]{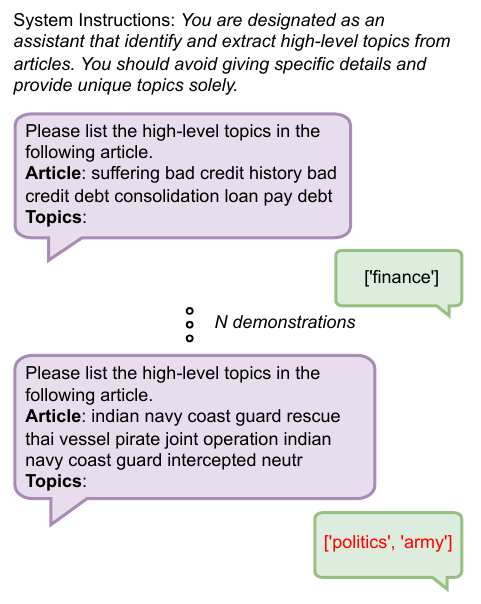} 
	\caption{\textsf{PromptTopic}'s \textit{topic generation}: \textit{N} turns of user input (\textcolor{user_input}{purple}) and assistant's sample answer (\textcolor{assistant_sample}{green}). ChatGPT's output answer is shown in \textcolor{generated_gpt}{red}.}
	\label{fig:prompt_generation}
\end{figure}



Figure \ref{fig:prompt_generation} displays the prompt setup used for topic generation using ChatGPT. The prompt input comprises $N$ demonstration examples, each with prompt inputs and their associated annotated answers. For LLaMA, which isn't instruction-trained, the instructional statements are omitted from the prompt.

To find the optimal value for the demonstration parameter $N$, we tested values of 2, 4, 6, and 8. Our results indicated that a setting of $N = 4$ produced the best topic generation performance, notably reducing errors in the offline LLaMA model. ChatGPT, with its larger parameter size, was less sensitive to changes in $N$. Table \ref{tab:topics_generation} presents topics generated by LLaMA for a randomly selected document, with $N$ ranging from 2 to 8. 
\begin{table*}[htbp]
    \centering
    \small
        \caption{Topics Generated by LLaMA for Demonstration Numbers (N) from 2 to 8, with Current Document-Related Topics Highlighted.}
    \begin{tabular}{p{7cm}p{2.3cm}p{2.3cm}p{2.3cm}p{1.2cm}}
        \toprule
        \textbf{Document} & \multicolumn{4}{c}{\textbf{N (Number of Demonstrations)}} \\
        \cmidrule{2-5}
         & 2 & 4 & 6 & 8 \\
        \midrule
        trailer talk week movie rite mechanic week opportunity & politics, army & \textbf{movies}, \textbf{trailers}, \textbf{mechanic} & politics, economy, finance & social \\
        \bottomrule
    \end{tabular}
    \label{tab:topics_generation}
\end{table*}

\subsection{Collapsing Overlapping Topics}
LLMs often generate overlapping topics for a document. For instance, topics such as “film”, and “actor” can be merged into one topic “film”. We introduced two approaches to collapse topics: \textit{Prompt-Based Matching (PBM)} and \textit{Word Similarity Matching (WSM)}.
 
\paragraph{Prompt-Based Matching (PBM).} We utilize prompts to group similar topics. The process involves sorting the unique topics based on their frequency counts and selecting a specified $K$ number of topics. Initially, we form a set $T_{n}$ containing the unique topics. Then, we create a subset $T_{n-1}$ by selecting the first $n-1$ topics from $T_n$. Next, we prompt LLMs to merge each topic $t_n$ from $T_n$ with a topic from $T_{n-1}$. If no merge is possible, we merge $t_n$ with the "\textit{Miscellaneous}" topic. This iterative process continues until $T_n$ contains only $K$ topics.

During experimentation, we encountered a challenge with datasets that had a large number of unique topics, exceeding the maximum token length allowed by LLMs. To overcome this, we employed a sliding window approach. We selected a window of size $M$ from the sorted unique topic set and performed the iterative topic grouping process. If LLMs successfully merged a topic with one of the $M$ topics, we merged them and restarted the iteration. If no merge occurred, we categorized the topic as "\textit{Miscellaneous}."

\paragraph{Word Similarity Matching (WSM).} This approach involves computing topic similarity and merging highly similar topics. We aggregate documents associated with each topic and compute a Class-based Term Frequency-Inverse Document Frequency (c-TF-IDF) representation, which captures word frequencies within topics while considering their importance across all topics. We select the top 20 words from the c-TF-IDF representation, retaining only relevant words closely related to the topic. Topic similarity is measured by counting the number of common words between the top words of each topic pair, normalized by word count. We merge the pair with the highest word similarity, create a new topic, and recalculate the similarity score based on the merged content. This process is iterated until we have $K$ unique topics remaining.

To reduce computation time for large datasets with numerous topics, we utilize the PBM model to compress the initial $n$ topics into a more concise set of topics, designated as $G$. This $G$ value exceeds the intended number of topics, $K$, yet remains substantially smaller than the initial number of topics, $n$. Subsequently, we will proceed to further condense the set of $G$ topics to the desired $K$ through the utilization of the WSM technique.

\subsection{Topic Representation Generation}
In order to evaluate the performance of our \textsf{PromptTopic} model, we utilized well-established topic model metrics. However, evaluating the topics required representing them as word mixtures. Since our model did not generate topic-word distributions directly, we employed c-TF-IDF scores to compute the most representative words for each cluster. Initially, we obtained the top 100 c-TF-IDF words for each topic. To further refine the representation, we used LLMs to filter these words down to the top 10 most representative words. This process allowed us to assess the quality and coherence of the generated topics based on their representative word mixtures.

\section{Experiment}
\begin{table}[t]
    \centering
    \caption{Dataset statistics: Size indicates the number of documents, Category indicates the number of categories, Text indicates the average length of documents.}
    \begin{tabular}{cccc}
       \hline
        \textbf{Dataset} & \textbf{Size} & \textbf{Category} & \textbf{Text} \\
        \hline
        20 NewsGroup & 16,309 & 20 & 185.37 \\
        \hline
        Yelp Reviews & 10,000 & - &131.03 \\
        \hline
        Twitter Tweet & 2,472 & 89 & 8.55 \\
        \hline
    \end{tabular}
    \label{tab:dataset_statistics}
\end{table}


\begin{table*}[ht]
\small
    \centering
    {
    \caption{Comparison of NPMI and Topic Diversity (TD) across three datasets. Cell color intensity indicates the score's ranking among all models, with the best performing model underlined.}
    \begin{tabular}{ccc|cc|cc}
       \hline
       & \multicolumn{2}{c}{\textbf{20 NewsGroup}} & 
       \multicolumn{2}{c}{\textbf{Yelp Reviews}} & 
       \multicolumn{2}{c}{\textbf{Twitter Tweet}} \\
        \cmidrule{2-7}
        \textbf{Model} & NPMI & TD & NPMI & TD  & NPMI & TD \\
                \hline
LDA& \cellcolor{green!20}-0.05& \cellcolor{blue!10}0.81& \cellcolor{green!30}-0.01& \cellcolor{blue!5}0.43& \cellcolor{green!15}-0.36& \cellcolor{blue!10}0.41\\ 
 \hline 
NMF& \cellcolor{green!35}0.04& \cellcolor{blue!5}0.63& \cellcolor{green!35}0.02& \cellcolor{blue!10}0.45& \cellcolor{green!20}-0.25& \cellcolor{blue!15}0.60\\ 
 \hline 
CTM& \cellcolor{green!30}-0.01& \cellcolor{blue!35}0.96& \cellcolor{green!20}-0.09& \cellcolor{blue!20}0.78& \cellcolor{green!40}0.03& \cellcolor{blue!35}0.96\\ 
 \hline 
TopClus& \cellcolor{green!10}-0.13& \cellcolor{blue!30}0.92& \cellcolor{green!15}-0.13& \cellcolor{blue!35}0.92& \cellcolor{green!10}-0.37& \cellcolor{blue!25}0.92\\ 
 \hline 
Cluster-Analysis& \cellcolor{green!25}-0.02& \cellcolor{blue!45}0.99& \cellcolor{green!25}-0.04& \cellcolor{blue!40}0.96& \cellcolor{green!5}-0.43& \cellcolor{blue!5}0.27\\ 
 \hline 
BERTopic& \cellcolor{green!45}0.10& \cellcolor{blue!40}0.97& \cellcolor{green!50}0.10& \cellcolor{blue!25}0.81& \cellcolor{green!50}0.05& \cellcolor{blue!45}0.98\\ 
 \hline 
\textsf{PromptTopic-PBM}(LLaMA)& \cellcolor{green!15}-0.12& \cellcolor{blue!40}0.97& \cellcolor{green!10}-0.24& \cellcolor{blue!50}0.99& \cellcolor{green!25}-0.14& \cellcolor{blue!20}0.91\\ 
 \hline 
\textsf{PromptTopic-PBM}(ChatGPT)& \cellcolor{green!5}-0.15& \cellcolor{blue!20}0.89& \cellcolor{green!5}-0.26& \cellcolor{blue!45}0.98& \cellcolor{green!35}-0.04& \cellcolor{blue!30}0.95\\ 
 \hline 
\textsf{PromptTopic-WSM}(LLaMA)& \cellcolor{green!40}0.05& \cellcolor{blue!25}0.91& \cellcolor{green!40}0.04& \cellcolor{blue!30}0.86& \cellcolor{green!45}0.04& \cellcolor{blue!40}0.97\\ 
 \hline 
\textsf{PromptTopic-WSM}(ChatGPT)& \cellcolor{green!35}0.04& \cellcolor{blue!15}0.84& \cellcolor{green!45}0.08& \cellcolor{blue!15}0.76& \cellcolor{green!30}-0.07& \cellcolor{blue!20}0.91\\ 
 \hline 
    \end{tabular}
    \label{tab:metrics_table}
    }
\end{table*}

\begin{table*}[!t]
\small
    \centering
    
        \caption{Qualitative evaluation of the topic-words representation in Twitter Tweet dataset. A subset of topics that occur frequently are selected. The related words belonging to the corresponding topic are highlighted in bold. }
    \begin{tabular}{ll|cccc|cc}
       \toprule
         \multirow{2}{*}{ \textbf{Dataset}} & \multirow{2}{*}{ \textbf{Topic}}  & \multicolumn{4}{c}{\textbf{\textsc{Baseline Model}}} & \multicolumn{2}{c}{\textbf{\textsc{\textsf{PromptTopic}}}}\\
        & & NMF & CTM &  Cluster-Analysis & BERTopic & WSM(LLaMA) & PBM(LLaMA) \\
        \hline

        \multirow{10}{*}{\rotatebox[origin=c]{90}{\textbf{Twitter Tweet}}} 
         & \multirow{5}{*}{Politics}
& \textbf{yemen} & \textbf{protest} & \textbf{ king} & \textbf{egypt} & \textbf{president} & \textbf{president} \\ 
& &\textbf{protest} & thousand & christina & \textbf{yemen} & \textbf{obama} & \textbf{protest} \\ 
& &\textbf{president} & \textbf{yemen} & \textbf{egypt } & journalist & judge & \textbf{protester} \\ 
& & aquarium & \textbf{government} & \textbf{ yemen} & \textbf{president} & \textbf{law} & \textbf{government} \\ 
& & \textbf{somali} & \textbf{yemeni} &sundance & \textbf{protest} & \textbf{lawsuit} & judge \\ 

\cline{2-8}
        & \multirow{5}{*}{Sports} 
&\textbf{superbowl} & fish & \textbf{superbowl } & commercial & \textbf{hockey} & \textbf{nba} \\ 
& &commercial & aquarium & \textbf{super }  & \textbf{superbowl} & sidney & \textbf{nfl} \\ 
& & \textbf{super} & bass & christina & \textbf{super} &\textbf{crosby} & \textbf{football} \\ 
& & \textbf{bowl} & \textbf{fishing} & egypt & \textbf{bowl} & \textbf{concussion} & \textbf{tennis} \\ 
& &ad & \textbf{tackle} & yemen & christina & \textbf{safety} & \textbf{hockey} \\ 
\hline

    \end{tabular}
    \label{tab:demo_table}
\end{table*}

\subsection{Experiment Settings}

\textbf{Datasets.}\label{sec:datasets} We evaluate \textsf{PromptTopic} and baselines on three commonly-used topic modeling datasets: \textit{20 NewsGroup }\cite{lang1995newsweeder},  \textit{Yelp Reviews} \cite{yelpdataset} and \textit{Twitter Tweet} \cite{qiang2022shorttexttm}. Table \ref{tab:dataset_statistics} provide a statistical summary of the datasets


\textbf{Baseline Models.} \textsf{PromptTopic} will be evaluated against six widely used topic modeling models: LDA~\cite{blei2003latent}, NMF~\cite{fevotte2011algorithms}, CTM~\cite{bianchi2020cross}, TopClus~\cite{meng2022topicdiscovery}, Cluster-Analysis~\cite{sia2020tired} and BERTopic~\cite{grootendorst2022bertopic}.


\textbf{PromptTopic Configurations.} In our experiments, we applied the \textbf{PromptTopic} model to two state-of-the-art LLMs: ChatGPT and LLaMA-13B \cite{touvron2023llama}. However, due to the limited parameter size and absence of instruction training in LLaMA, we simplified the prompt format and reduced the number of demonstration examples. By adopting this approach, we ensured that the performance of LLaMA was comparable to that of ChatGPT.
Note that for all models, we preprocess the datasets using the OCTIS package~\cite{terragni2020octis}, which involved removing punctuation, stopwords, and performing lemmatization (except the Twitter Tweet dataset).

\textbf{Setting Optimal Value of Parameter $G$}. To determine the optimal value of parameter $G$ across three datasets introduced in Datasets. We conducted an empirical investigation encompassing $G$ values of 200, 400, 600, and 800. We evaluated $G$ based on two main criteria: the time for topic collapsing and quantitative assessments using established metrics from the Quantitative Evaluation Section~\ref{sec:evaluation_metrics}, including topic coherence (NPMI \cite{bouma2009tc}) and topic diversity (TD \cite{dieng2020etm}).
After conducting our assessment, we determined that the optimal value for parameter $G$ was 400 for both the 20 NewsGroup and Twitter Tweet datasets, while for the Yelp Reviews dataset, a value of 200 yielded the best results. Consequently, we selected the most effective value of $G$ for the subsequent experiments.

\subsection{Topic Evaluation}\label{sec:evaluation_metrics}

\textbf{Quantitative Evaluation.} Evaluations of topic modeling often use two well-established metrics: topic coherence and topic diversity. Topic coherence gauges the extent to which the words within a topic are related, forming a coherent group. It is typically calculated using statistics and probabilities drawn from the reference corpus, focusing specifically on the context of the words. In our experiments, we employed Normalized Pointwise Mutual Information (NPMI) \cite{bouma2009tc} as our measure of topic coherence. A higher NPMI score signifies better coherence, with a perfect correlation being represented by a score of 1.

Conversely, topic diversity \cite{dieng2020etm} evaluates the proportion of unique words across all topic representations. The diversity score ranges from 0 to 1, where a score of 0 indicates repetitive topics, and a score of 1 indicates diverse topics. This metric is crucial for ensuring that a topic model covers a wide range of themes without overemphasizing any particular topic. Using these two metrics together provides insights into the effectiveness of topic modeling algorithms in identifying both coherent and diverse topics.

In our evaluation process, we empirically selected the number of topics ($K$) for each dataset. $K$ is set to 40, 20, 20 for 20NewsGroup, Yelp Reviews, and Twitter Tweet Datasets, respectively. Table \ref{tab:metrics_table} shows NPMI and topic diversity scores for different topic models on three datasets. Our findings indicate that \textsf{PromptTopic-WSM} consistently outperforms the majority of baseline topic models across all datasets, as evidenced by both metrics. Notably, compared to the best-performing baseline model, BERTopic, the performance of \textsf{PromptTopic} remains comparable.

Remarkably, LLaMA-13b, functioning as a standalone offline model, exhibits significantly fewer parameters while yielding results of comparable quality to ChatGPT. Their coherency levels closely align, although LLaMA-13b demonstrates a propensity for generating more diverse topics, with a higher TD score. Consequently, in the subsequent phases of qualitative and human evaluation, we shall opt for LLaMA as our preferred Language Model.

\textbf{Human Evaluation.} Low NPMI scores don't necessarily reflect poor topic quality, as they have been found to show a weak correlation with human ratings \cite{hoyle2021automated}. To gain a deeper understanding, we decided to conduct a manual evaluation.

Our assessment of topic quality relied on the word intrusion task, following \cite{chang2009reading}. In this task, we presented participants with a list of five words. Four of these words were from the prominent words of a single topic generated by the model, while the fifth, known as the 'intruder', was randomly chosen from a different topic. The word intrusion test aims to determine if the five words collectively represent a clear and distinct topic, making it easy to identify the intruder.

In our study, we compared topic words generated by PromptTopic with those from the leading baseline model, BERTopic. We then report the intrusion task accuracy results for BERTopic, PromptTopic-PBM (LLaMA), and PromptTopic-WSM (LLaMA) across all topics in the three datasets. Each topic was evaluated by two annotators.

\begin{figure}[t]
	\centering
	\includegraphics[width=0.5\textwidth]{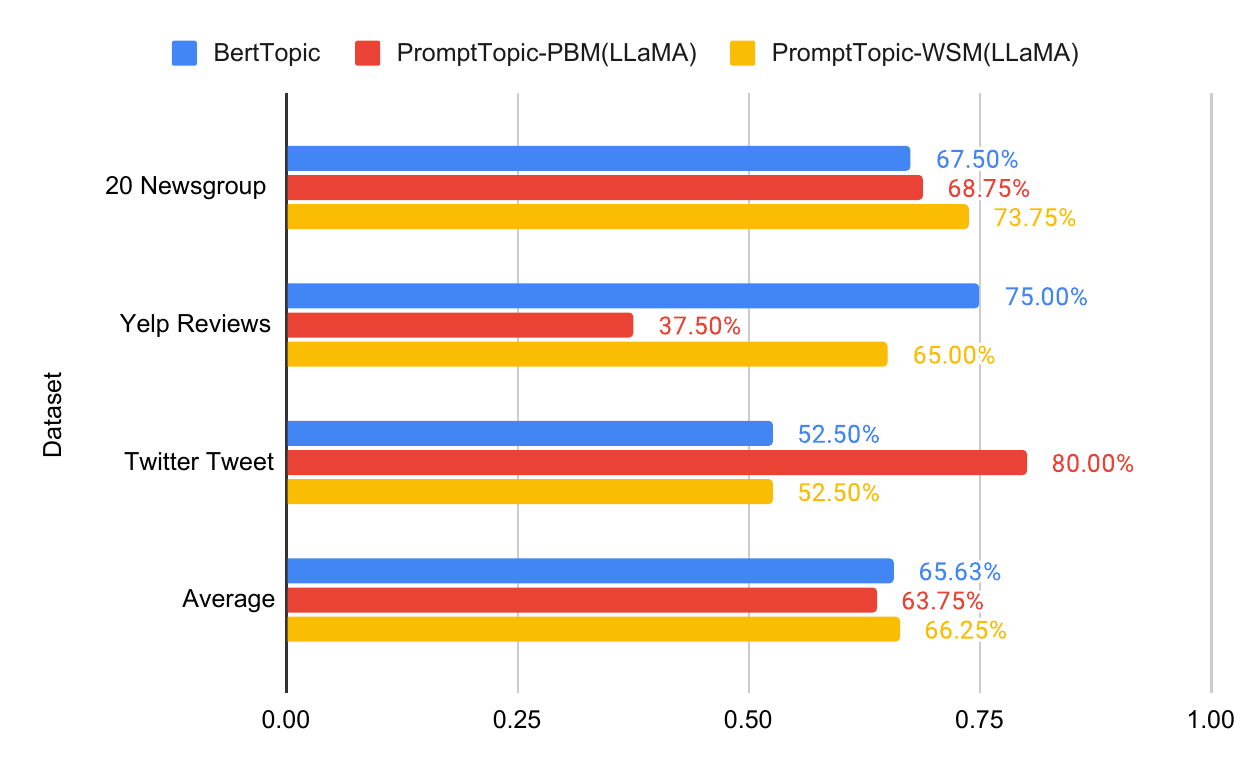} 
	\caption{Word Intrusion Study Results for 40, 20, and 20 Topics Across 20 NewsGroup, Yelp Reviews, and Twitter Tweet Datasets, generated by BERTopic, \textsf{PromptTopic-PBM}(LLaMA) and \textsf{PromptTopic-WSM}(LLaMA) Models. Average reflects the overall accuracy across all three datasets.}
	\label{fig:human_evaluation}
\end{figure}

Figure \ref{fig:human_evaluation} presents word intrusion task results for three models across three diverse datasets. We observed a high level of consistency in word intrusion accuracy, with an average score of around 65\% across all datasets. This suggests that the generated topic words maintain consistent high quality. However, it's essential to note that model performance is significantly influenced by dataset characteristics. Regarding \textsf{PromptTopic-WSM} and BERTopic, their performance is particularly strong when dealing with lengthy textual datasets like 20 NewsGroup and Yelp Reviews. However, they exhibit reduced performance when handling short text data, as seen in Twitter Tweets.
In contrast, \textsf{PromptTopic-PBM} presents a unique scenario. It performs suboptimally in the Yelp Reviews dataset due to the dataset's strong focus on food-related content. This concentration results in \textsf{PromptTopic-PBM} generating highly specific topics, leading to a topic diversity score of 0.99. However, the disjointed occurrence of these specific food-related words within the same document affects coherence negatively. On the other hand, \textsf{PromptTopic-PBM} excels in the Twitter Tweet dataset, achieving an impressive word intrusion task accuracy of 0.80, surpassing the performance of the other models. This highlights \textsf{PromptTopic-PBM}'s effectiveness in handling short text data, leveraging the power of Language Model Models (LLMs) to extract relevant topics.

\textbf{Qualitative Evaluation.} We randomly selected commonly occurring topics from all datasets and performed a manual matching procedure to determine the most relevant topic generated by each model.
Table \ref{tab:demo_table} illustrates the top five words associated with each topic, generated by employing the LLaMA method for the \textsf{PromptTopic}. To enhance page space efficiency without compromising clarity, we have selectively showcased the Twitter Tweet dataset and restricted the presentation to four fundamental models, notable for their strong topical coherence across the majority of datasets, as delineated in the Quantitative Evaluation Section~\ref{sec:evaluation_metrics}. Despite BERTopic's higher NPMI score, the manual assessment reveals that \textsf{PromptTopic-PBM} demonstrates comparable topic representation. BERTopic falls short in the short text such as the Sports topic in Twitter Tweet dataset, providing only three relevant words, with only `\textit{superbowl}' being informative. In contrast,  \textsf{PromptTopic-PBM} generates informative words like `\textit{nba}', `\textit{nfl}', `\textit{football}', `\textit{tennis}', and `\textit{hockey}' without any overlap. The observed enhancement of \textsf{PromptTopic-PBM} performance in short text datasets aligns with human evaluation findings, which can be attributed to to LLMs' robust language comprehension and vast knowledge base.

\section{Limitations}
When dealing with large datasets, the usage of LLMs for topic generation can be resource-intensive. LLMs like LLaMA require GPU devices with significant memory capacity. In our experiments, we employed the PBM method to collapse topics by prompting the LLM to merge a topic with a list of topics solely based on topic names. However, this approach lacks context and may result in the merging of unrelated topics. Furthermore, in datasets characterized by a substantial number of topics, the need for batch-wise merging in PBM and the assistance of PBM in WSM becomes imperative.

\section{Conclusion}

In this paper, we introduce \textsf{PromptTopic}, a groundbreaking approach to topic modeling utilizing LLMs. Our innovative method stands out by harnessing the power of LLMs to discern semantic structures both at the token and sentence levels. This ensures the generation of not just coherent, but also diverse topics. Through rigorous evaluations of diverse datasets, we not only validate the robustness of our approach but also highlight its superior capabilities. When compared to leading contemporary methods, \textsf{PromptTopic} not only matches them in terms of automatic metrics but notably surpasses them in producing more meaningful topics upon qualitative assessment. This underscores the significant contribution of our work to the field. In future projects, we'll investigate ways to enhance batch-wise merging in PBM and utilize prompt-engineering methods for better topic modeling.

\balance
\bibliography{ref}
\bibliographystyle{IEEEtran}

\end{document}